\documentclass{article}

\PassOptionsToPackage{numbers, compress}{natbib}

\usepackage{subfig}
\usepackage[utf8]{inputenc} % allow utf-8 input
\usepackage[T1]{fontenc}    % use 8-bit T1 fonts
\usepackage{url}            % simple URL typesetting
\usepackage{booktabs}       % professional-quality tables
\usepackage{amsfonts}       % blackboard math symbols
\usepackage{nicefrac}       % compact symbols for 1/2, etc.
\usepackage{microtype}      % microtypography
\usepackage{xcolor}    
\usepackage{color, colortbl}% colors
\usepackage{algorithm}
\usepackage{algorithmic}
\usepackage[pdftex]{graphicx}
\usepackage{appendix}
\usepackage{multirow}
\usepackage{amsmath}
\usepackage{comment}
\usepackage[colorlinks]{hyperref} 
\usepackage{xcolor}
\usepackage{xspace}
\usepackage{enumitem}
\usepackage[normalem]{ulem}

\definecolor{LightCyan}{rgb}{0.88,1,1}

\hypersetup{colorlinks,allcolors=[rgb]{0.6,0.15,0.00098}}

\usepackage[final]{neurips_2024}

\usepackage[utf8]{inputenc} % allow utf-8 input
\usepackage[T1]{fontenc}    % use 8-bit T1 fonts
\usepackage{hyperref}       % hyperlinks
\usepackage{url}            % simple URL typesetting
\usepackage{booktabs}       % professional-quality tables
\usepackage{amsfonts}       % blackboard math symbols
\usepackage{nicefrac}       % compact symbols for 1/2, etc.
\usepackage{microtype}      % microtypography
\usepackage{xcolor}         % colors

\title{Pixel is a Barrier: Diffusion Models Are More Adversarially Robust Than We Think}

\author{%
  Haotian Xue \\
  Georgia Institute of Technology\\
  htxue.ai@gatech.edu\\
  \And
   Yongxin Chen\\
  Georgia Institute of Technology\\
  yongchen@gatech.edu\\
}

\begin{document}

\maketitle

\begin{abstract}

% Adversarial examples for diffusion models are widely used as solutions for safety concerns. By adding adversarial perturbations to personal images, attackers can not edit or imitate them easily. However, it is essential to note that all these protections target the latent diffusion model (LDMs), the adversarial examples for diffusion models in the pixel space (PDMs) are largely overlooked. This may mislead us to think that the diffusion models are vulnerable to adversarial attacks like most deep models. In this paper, we show novel findings that: even though gradient-based white-box attacks can be used to attack the LDMs, they fail to attack PDMs. This finding is supported by extensive experiments of almost a wide range of attacking methods on various PDMs and LDMs with different model structures, meaning diffusion models are much more robust against adversarial attacks. Moreover, we find that PDMs can be used as an off-the-shelf purifier to effectively remove the adversarial patterns that were generated on LDMs to protect the images, which means that most protection methods nowadays, to some extent, cannot protect our images from malicious attacks. We hope that our insights will inspire the community to rethink the adversarial samples for diffusion models as protection methods and move forward to more effective protection. Codes are available in \url{https://github.com/xavihart/PDM-Pure}.

Diffusion models have demonstrated an impressive capability to edit or imitate images, which has raised concerns regarding the safeguarding of intellectual property. To address these concerns, the adoption of adversarial attacks, which introduce adversarial perturbations into protected images, has proven successful. Consequently, diffusion models, like many other deep network models, are believed to be susceptible to adversarial attacks. However, in this work, we draw attention to an important oversight in existing research, as all previous studies have focused solely on attacking latent diffusion models (LDMs), neglecting adversarial examples for diffusion models in the pixel space (PDMs). Through extensive experiments, we demonstrate that nearly all existing adversarial attack methods designed for LDMs fail when applied to PDMs. We attribute the vulnerability of LDMs to their encoders, indicating that diffusion models exhibit strong robustness against adversarial attacks. Building upon this insight, we propose utilizing PDMs as an off-the-shelf purifier to effectively eliminate adversarial patterns generated by LDMs, thereby maintaining the integrity of images. Notably, we highlight that most existing protection methods can be easily bypassed using PDM-based purification. We hope our findings prompt a reevaluation of adversarial samples for diffusion models as potential protection methods. Codes are available in \url{https://github.com/xavihart/PDM-Pure}.

% Generative Diffusion Models excel at generating high-quality images however, they can cause safety issues by maliciously editing or mimicking 

\end{abstract}

\section{Introduction}

Generative diffusion models (DMs)~\citep{ddpm,song2020score,ldm} have achieved great success in generating images with high fidelity. However, this remarkable generative capability of diffusion models is accompanied by safety concerns~\cite{zhang2023text}, especially on the unauthorized editing or imitation of personal images such as portraits or individual artworks~\cite{andersen2023,setty2023}.
Recent works~\cite{advdm, glaze,salman2023raising, sdsattack, mist-v2, chen2024smoothattack,ahn2024imperceptible, metacloak} show that adversarial samples (adv-samples) for diffusion models can be applied as a protection against malicious editing. Small perturbations generated by conventional methods in adversarial machine learning~\citep{pgd,goodfellow2014fgsm} can effectively fool popular diffusion models such as Stable Diffusion~\cite{ldm} to produce chaotic results when an imitation attempt is made. However, a significantly overlooked aspect is that all the existing works focus on latent diffusion models (LDMs) and the pixel-space diffusion models (PDMs) are not studied. For LDMs, perturbations are not directly introduced to the input of the diffusion models. Instead, they are applied externally and propagated through an encoder. It has been shown that the encoder-decoder of LDMs is vulnerable to adversarial perturbations ~\cite{zhang2023robustness,sdsattack}, which means that the adv-samples for LDMs have a very different mechanism compared with the adv-samples for PDMs. 
Moreover, some existing works~\cite{liang2023mist, salman2023raising} show that combining encoder-specific loss can enhance the adversary, ~\cite{sdsattack} further demonstrating that 
% the gradient of the denoising process is weak and unstable, and 
the encoder is the bottleneck for attacking LDMs. Building upon this observation, in this paper, we draw attention to
rethink existing adversarial attack methods for diffusion models:

\begin{center}
    \textit{Can we generate adversarial examples for PDMs as we did for LDMs?}
\end{center}

% \chen{combine this with the first paragraph before raising our research question} 

% Instead of attacking the diffusion process itself, current adversarial examples for Latent Diffusion Models heavily rely on attacking the encoder: Glaze~\cite{glaze} is built on minimizing the distance between the attacked image and the target image in the latent space defined by the encoder. Additionally, Mist~\cite{liang2023mist} demonstrates the significance of combining the textural loss derived from the encoder to generate better adversarial samples. Moreover, SDS-attack \cite{sdsattack} further investigates that, the gradient of denoising process is weak and unstable, and the real bottleneck of attacking an LDM is attacking the encoder.
%

We address this question by systematically investigating adv-samples for PDMs. We conduct experiments on various LDMs or PDMs with different network architectures (e.g. U-Net~\cite{ddpm} or Transformer~\cite{dit}), different training datasets, and different input resolutions (e.g. 64, 256, 512). Through extensive experiments, we demonstrate that all the existing methods we tested ~\citep{liang2023mist,mist-v2, glaze, sdsattack, chen2024smoothattack, salman2023raising, advdm}, targeting to attack LDMs, fail to generate effective adv-samples for PDMs. This implies that PDMs are more adversarial robust than we think. 

% \chen{this is confusing More importantly, it means that the previous adversarial examples for diffusion models (AdvDM) are, in fact, one special case of adv-samples for LDMs (AdvLDM) only.}

Building on this insight that PDMs are strongly robust against adversarial perturbations, we further propose PDM-Pure, a universal purifier that can effectively remove the protective perturbations of different scales (e.g. Mist-v2~\cite{mist-v2} and Glaze~\cite{glaze}) based on PDMs trained on large datasets. Through extensive experiments, we demonstrate that PDM-Pure achieves way better performance than all baseline methods.

To summarize, the pixel is a barrier to adversarial attack; the diffusion process in the pixel space makes PDMs much more robust than LDMs. This property of PDMs also makes real protection against the misusage of diffusion models difficult since all the existing protections can be easily purified using a strong PDM. Our contributions are listed below.

\begin{enumerate}[parsep=0pt,topsep=0pt,leftmargin=12pt]
    \item We observe that most existing works on adversarial examples for protection focus on LDMs. Adversarial attacks against PDMs are \textbf{largely overlooked} in this field. 
    \item We fill in the gap in the literature by conducting extensive experiments on various LDMs and PDMs. We discover that all the existing methods \textbf{fail} to attack the PDMs, indicating that PDMs are much more adversarially robust than LDMs.
    \item Based on this novel insight, we propose a simple yet effective framework termed PDM-Pure that applies strong PDMs as \textbf{a universal purifier} to remove attack-agnostic adversarial perturbations, easily bypassing almost all existing protective methods. 

\end{enumerate}

% we made an observation that most of the works focus on LDM, PDMs are not studied

% we test it and find that PDM is much more robust than LDMs 

% PDM-Pure 

\begin{figure}[t]
  \centering
\includegraphics[width=0.99\linewidth]{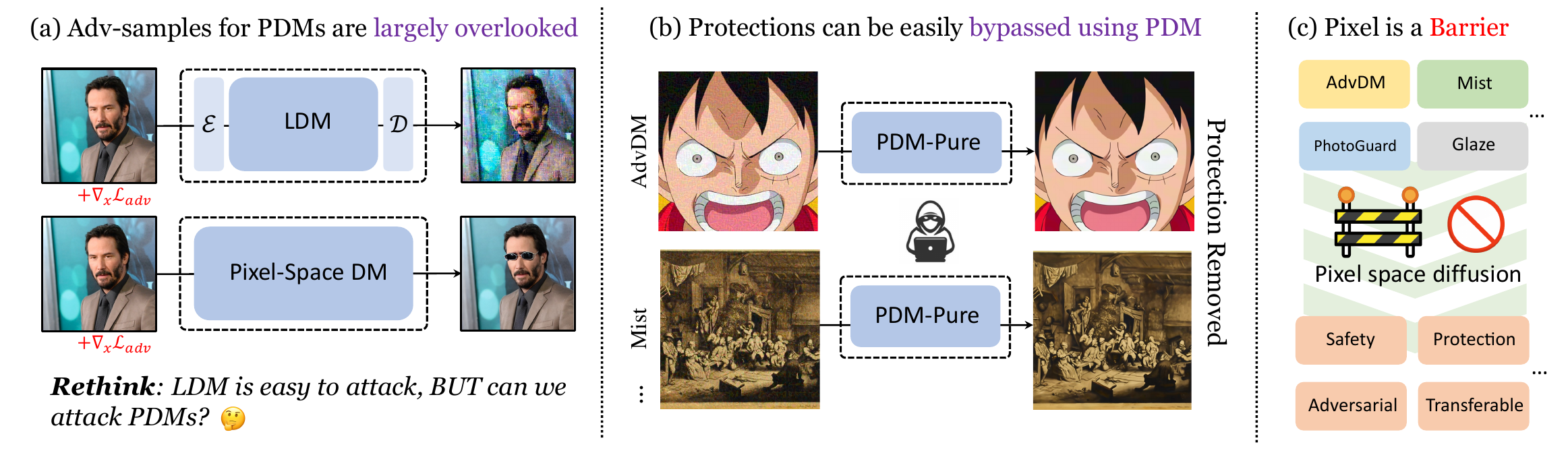}
  \vspace{-5pt}
  \caption{\textbf{Pixel is a Barrier for Attacking DMs}: (a) Pixel-based diffusion models are harder to attack using white-box attacks like project-gradient-descent than diffusion models in the latent space. (b) Strong PDM can be used as a universal purifier to effectively remove the protective perturbation generated by existing protection methods. (c) Pixel is a barrier and the pixel-space diffusion model is quite robust, and we cannot achieve real safety and protection if pixel-space diffusion is not attacked. }

  \label{fig:teaser}
  \vspace{-0.4cm}
\end{figure}

% background

% issues

% what we find

%

\section{Related Works}

\paragraph{Safety Issues in Diffusion Models}
The impressive generative capability of the diffusion models has raised numerous safety issues~\cite{zhang2023text,setty2023,andersen2023}. As a result, there has been a growing interest in preventing DMs from being abused. Some of the existing works focus on the protection of intellectual property of diffusion models by applying watermarks~\citep{zhao2023recipe, peng2023protecting, cui2023diffusionshield} and some of them are on concept removal to prevent the DMs from generating NSFW images~\citep{heng2023continual,zhang2023forget,gandikota2023unified}. In the era of generative models, caution should be taken to guarantee safe and responsible applications of these models.

\paragraph{Adversarial Examples for DMs} Adversarial samples~\cite{goodfellow2014fgsm, carlini2017towards, glaze} are clean samples perturbed by an imperceptible small noise that can fool the deep neural networks into making wrong decisions. Under the white-box settings, gradient-based methods are widely used to generate adv-samples. Among them, the projected gradient descent (PGD) algorithm~\cite{pgd} is one of the most effective methods. Recent works~\citep{advdm, salman2023raising} show that it is also easy to find adv-samples for diffusion models (AdvDM): with a proper loss to attack the denoising process, the perturbed image can fool the diffusion models to generate chaotic images when operating diffusion-based mimicry. Furthermore, many improved algorithms~\cite{mist-v2,chen2024smoothattack,sdsattack} have been proposed to generate better AdvDM samples. However, to our best knowledge, all the AdvDM methods listed above are used on LDMs, and those for the PDMs are rarely explored.

\paragraph{Adversarial Perturbation as Protection} Adversarial perturbation against DMs turns out to be an effective method to safeguard images against unauthorized editing~\cite{advdm, glaze,salman2023raising, sdsattack, mist-v2, chen2024smoothattack,ahn2024imperceptible, metacloak}. It has found applications (e.g., Glaze~\cite{glaze} and Mist~\cite{mist-v2, liang2023mist}) for individual artists to protect their creations. SDS-attack~\cite{sdsattack} further investigates the mechanism behind the attack and proposes some tools to make the protection more effective. However, they are limited to protecting LDMs only.
% without further investigating whether they can work for more general PDMs. 
In addition, some works~\cite{zhao2023can, sandoval2023jpeg} find that these protective perturbations can be purified. For instance, GrIDPure~\cite{zhao2023can} find that DiffPure~\cite{nie2022diffusion} can be used to purify the adversarial patterns, but they did not realize that the reason behind this is the robustness of PDMs.

\begin{figure}[t]
  \centering
\includegraphics[width=0.99\linewidth]{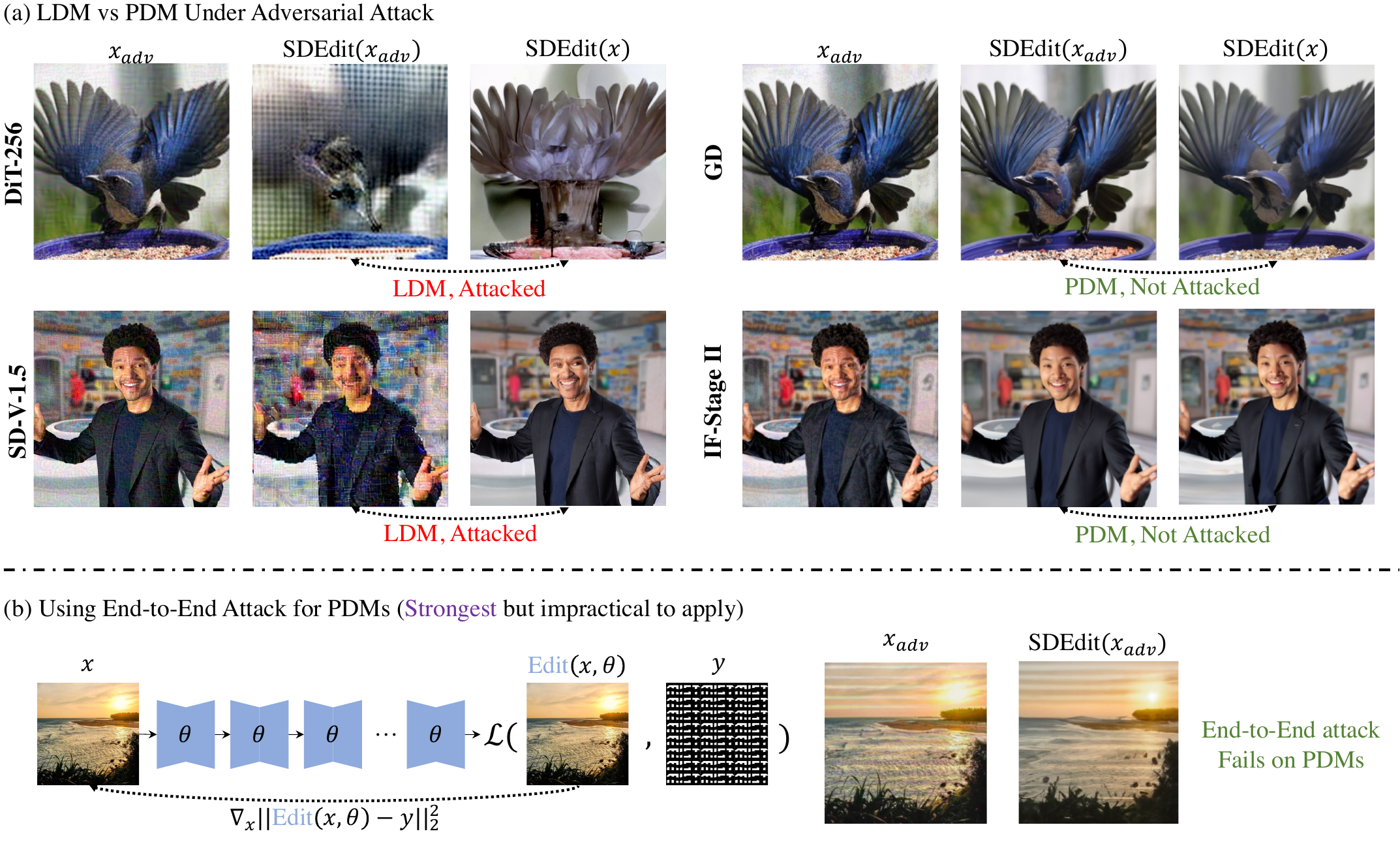}
  \caption{\textbf{PDMs Cannot be Attacked as LDMs}: (a) LDMs can be easily fooled but PDMs cannot be. (b) Even End-to-End attack does not work on PDMs. (Best viewed with zoom-in)}

  \label{fig:attack_various_models}
  \vspace{-0.4cm}
\end{figure}

\section{Preliminaries}

\paragraph{Generative Diffusion Models}

The generative diffusion model~
\cite{ddpm, song2020score} is one type of generative model, and it has demonstrated remarkable generative capability in numerous fields such as image~\cite{ldm, balaji2022ediffi}, 3D~\cite{poole2022dreamfusion, lin2022magic3d}, video~\cite{vdm,makeavideo}, story~\cite{pan2022story, rahman2023make} and music~\cite{musicdiff, huang2023noise2music} generation. Diffusion models, like other generative models, are parametrized models $p_{\theta}(\hat{x}_0)$ that can estimate an unknown distribution $q(x_0)$. For image generation tasks, $q(x_0)$ is the distribution of real images.

There are two processes involved in a diffusion model, a forward diffusion process and a reverse denoising process. The forward diffusion process progressively injects noise into the clean image, and the $t$-th step diffusion is formulated as $q(x_t \mid x_{t-1} ) = \mathcal{N} (x_t; \sqrt{1 - \beta_t}x_{t-1}, \, \beta_t \mathbf{I})$. Accumulating the noise, we have $    q_t(x_t \mid x_0 ) = \mathcal{N} (x_t; \sqrt{\bar{\alpha}_t} \, x_{t-1}, \, (1-\bar{\alpha}_t) \mathbf{I})$. Here $\beta_t$ growing from $0$ to $1$ are pre-defined values,  $\alpha_t = 1-\beta_t$, and $\bar{\alpha}_t = \Pi_{s=1}^{t} \alpha_s$. Finally, $x_T$ will become approximately an isotropic Gaussian random variable when $\bar{\alpha}_t \rightarrow 0$. 

Reversely, $p_{\theta}(\hat{x}_{t-1}|\hat{x}_{t})$ can generate samples from Gaussian $\hat{x}_T \sim\mathcal{N} (0, \textbf{I})$, where $p_{\theta}$ be re-parameterized by learning a noise estimator $\epsilon_{\theta}$, the training loss is $\mathbb{E}_{t, x_0, \epsilon}[\lambda(t)\|\epsilon_{\theta}(x_t,t) - \epsilon \|^2]$ weighted by $\lambda(t)$, where $\epsilon$ is the noise used to diffuse $x_0$ following $q_t(x_t|x_0)$. Finally, by iteratively applying $p_{\theta}(\hat{x}_{t-1}|\hat{x}_{t})$, we can sample realistic images following $p_{\theta}(\hat{x}_0)$.

Since the above diffusion process operates directly in the pixel space, we call such diffusion models Pixel-Space Diffusion Models (PDMs). Another popular choice is to move the diffusion process into the latent space to make it more scalable, resulting in the Latent Diffusion Models (LDMs)~\cite{ldm}. More specifically, LDMs first use an encoder $\mathcal{E}_{\phi}$ parameterized by $\phi$ to encode $x_0$ into a latent variable $z_0 = \mathcal{E}_{\phi}(x_0)$. The denoising diffusion process is the same as PDMs. At the end of the denoising process, $\hat{z}_0$ can be projected back to the pixel space using decoder $\mathcal{D}_{\psi}$ parameterized by $\psi$ as $\hat{x}_0 = \mathcal{D}_{\psi}(\hat{z}_0)$.

% \chen{breifly introduce SDEdit?}

\begin{table}[t]
 
  \label{tab:my_label}
 \resizebox{\textwidth}{!}{%
  \centering
  \begin{tabular}{c|ccc|ccc|ccc|ccc|c}
  \toprule
    \multicolumn{1}{c}{Models} & 
    \multicolumn{3}{c}{\textbf{FID-score}$\uparrow$} & \multicolumn{3}{c}{\textbf{SSIM} $\downarrow$} & \multicolumn{3}{c}{\textbf{LPIPS} $\uparrow$} & \multicolumn{3}{c}{\textbf{IA-Score} $\downarrow$}  & \textbf{Type} \\
    \midrule
  $\delta=4/255$ & Clean & Adv & $\Delta$  & Clean & Adv & $\Delta$  & Clean & Adv & $\Delta$ & Clean & Adv & $\Delta$ &   \\  

    \midrule
DiT-256 & 131 & 167  & {\color{red}+36}  & 0.37 & 0.35 & {\color{red}-0.02} & 0.44 & 0.54 & {\color{red}+0.10} & 0.74 & 0.70 & {\color{red}-0.04} & LDM  \\
SD-V-1.4 & 44 & 114  & {\color{red}+70}  & 0.68 & 0.55 & {\color{red}-0.13}  & 0.22 & 0.46 & {\color{red}+0.24}  & 0.92 & 0.84 & {\color{red}-0.08}  & LDM  \\
SD-V-1.5 & 45 & 113 & {\color{red}+68}  & 0.73 & 0.59 & {\color{red}-0.14} & 0.20 & 0.38 & {\color{red}+0.138} & 0.94 & 0.89 & {\color{red}-0.05} & LDM  \\
GD-ImageNet & 109 & 109  & +0  & 0.66 & 0.66 & -0.00 & 0.21 & 0.21 & +0.00 & 0.90 & 0.90 & -0.00 & PDM  \\
IF-I & 186 & 187  & +1  & 0.59 & 0.58 & -0.01 & 0.14 & 0.14 & +0.00 & 0.86 & 0.86 & -0.00 & PDM  \\
IF-II & 85 & 87  & +2 & 0.84 & 0.84 & -0.00 & 0.15 & 0.15 & +0.00 & 0.91 & 0.91 & -0.00 & PDM   \\

\midrule
  $\delta=8/255$ & Clean & Adv & $\Delta$  & Clean & Adv & $\Delta$  & Clean & Adv & $\Delta$ & Clean & Adv & $\Delta$ &   \\  

    \midrule
DiT-256 &131 & 186  &{\color{red}+55}  & 0.37 & 0.31 & {\color{red}-0.06} & 0.44 & 0.63 & {\color{red}+0.19} & 0.74 & 0.66 & {\color{red}-0.08} & LDM  \\
SD-V-1.4 & 44 & 178  & {\color{red}+134}  & 0.68 & 0.44 & {\color{red}-0.24}  & 0.22 & 0.60 & {\color{red}+0.38}  & 0.92 & 0.78 & {\color{red}-0.14}  & LDM  \\
SD-V-1.5 & 45 & 179  & {\color{red}+134}  & 0.73 & 0.49 & {\color{red}-0.24} & 0.20 & 0.51 & {\color{red}+0.31} & 0.94 & 0.84 & {\color{red}-0.10} & LDM  \\
GD-ImageNet & 109 & 110  & +1  & 0.66 & 0.64 & -0.02 & 0.21 & 0.22 & +0.01 & 0.90 & 0.90 & -0.00 & PDM  \\
IF-I & 186 & 188  & +2  & 0.59 & 0.59 & -0.00 & 0.14 & 0.14 & +0.00 & 0.86 & 0.86 & +0.00 & PDM  \\
IF-II & 85 & 82   & -3  & 0.84 & 0.83 & -0.01 & 0.15 & 0.16 & +0.01 & 0.91 & 0.92 & +0.01 & PDM   \\

\midrule
  $\delta=16/255$ & clean & adv & $\Delta$  & clean & adv & $\Delta$  & clean & adv & $\Delta$ & clean & adv & $\Delta$ &   \\  

  \midrule
DiT-256 & 131 & 220  & {\color{red}+89}  & 0.37 & 0.26 & {\color{red}-0.11} & 0.44 & 0.70 & {\color{red}+0.26} & 0.74 & 0.63 & {\color{red}-0.11} & LDM  \\
SD-V-1.4 & 44 & 225  & {\color{red}+181}  & 0.68 & 0.34 & {\color{red}-0.34}  & 0.22 & 0.68 & {\color{red}+0.46}  & 0.92 & 0.72 & {\color{red}-0.20}  & LDM  \\
SD-V-1.5 & 45 & 226  & {\color{red}+181}  & 0.73 & 0.37 & {\color{red}-0.36} & 0.20 & 0.62 & {\color{red}+0.42} & 0.94 & 0.78 & {\color{red}-0.16} & LDM  \\
GD-ImageNet & 109 & 110  & +1  & 0.66 & 0.57 & -0.09 & 0.21 & 0.26 & +0.05 & 0.90 & 0.89 & -0.01 & PDM  \\
IF-I & 186 & 188  & +2  & 0.59 & 0.58 & -0.01 & 0.14 & 0.15 & +0.01 & 0.86 & 0.87 & +0.01 & PDM  \\
IF-II &85 & 86  & +1  & 0.84 & 0.76 & -0.08 & 0.15 & 0.21 & +0.06 & 0.91 & 0.95 & +0.04 & PDM   \\

\bottomrule
  \end{tabular}
}
\vspace{10pt}
 \caption{
  \textbf{Quantiative Measurement of PGD-based Adv-Attacks for LDMs and PDMs}: gradient-based diffusion attacks can attack LDMs effectively, making the difference $\Delta$ across all evaluation metrics between edited clean image and edited adversarial image large, which means the quality of edited images drops dramatically (in red). However, the PDMs are not affected much by the crafted adversarial perturbations, showing small $\Delta$ before and after the attacks.
  }
  \label{quant_protect}
  \vspace{-20pt}
\end{table}

\paragraph{Adversarial Examples for Diffusion Models} 
Recent works~\cite{salman2023raising, advdm} find that adding small perturbations to clean images will make the diffusion models perform badly in noise prediction, and further generate chaotic results in tasks like image editing and customized generation. The adversarial perturbations for LDMs can be generated by optimizing the Monte-Carlo-based adversarial loss:

\vspace{-0.4cm}
\begin{equation}\label{semantic_loss}
        \mathcal{L}_{adv}(x) = \mathbb{E}_{t, \epsilon} \mathbb{E}_{z_t \sim q_t(\mathcal{E}_{\phi}(x))}\|\epsilon_{\theta}(z_t, t) -\epsilon \|_2^2.
\end{equation}
\vspace{-0.4cm}

Other encoder-based losses~\cite{glaze, liang2023mist, mist-v2, sdsattack} further enhance the attack to make it more effective. With the carefully designed adversarial loss, we can run Projected Gradient Descent (PGD)~\cite{pgd} with $\ell_{\infty}$ budget $\delta$ to generate adversarial perturbations: 
% \chen{make it clear that $x$ is the clean image}
% \chen{use $x^k$? also $B_\infty(x^k, \delta)$?} \haotian{should be $B_\infty(x, \delta)$, since the budget is computed on the clean sample}

\vspace{-0.4cm}
\begin{equation}\label{pgd_update}    x^{k+1} = \mathcal{P}_{B_\infty(x^0, \delta)} \left[ x^{k} + \eta\, \text{sign}\nabla_{x^k}\mathcal{L}_{adv}(x^k) \right]
\end{equation}

 In the above equation, $\mathcal{P}_{B_\infty(x^0, \delta)}(\cdot)$ is the projection operator on the $\ell_\infty$ ball, where $x^0$ is the clean image to be perturbed. We use superscript $x^k$ to represent the iterations of the PGD and subscript $x_t$ for the diffusion steps.

\begin{figure}[t]
  \centering
\includegraphics[width=0.8\linewidth]{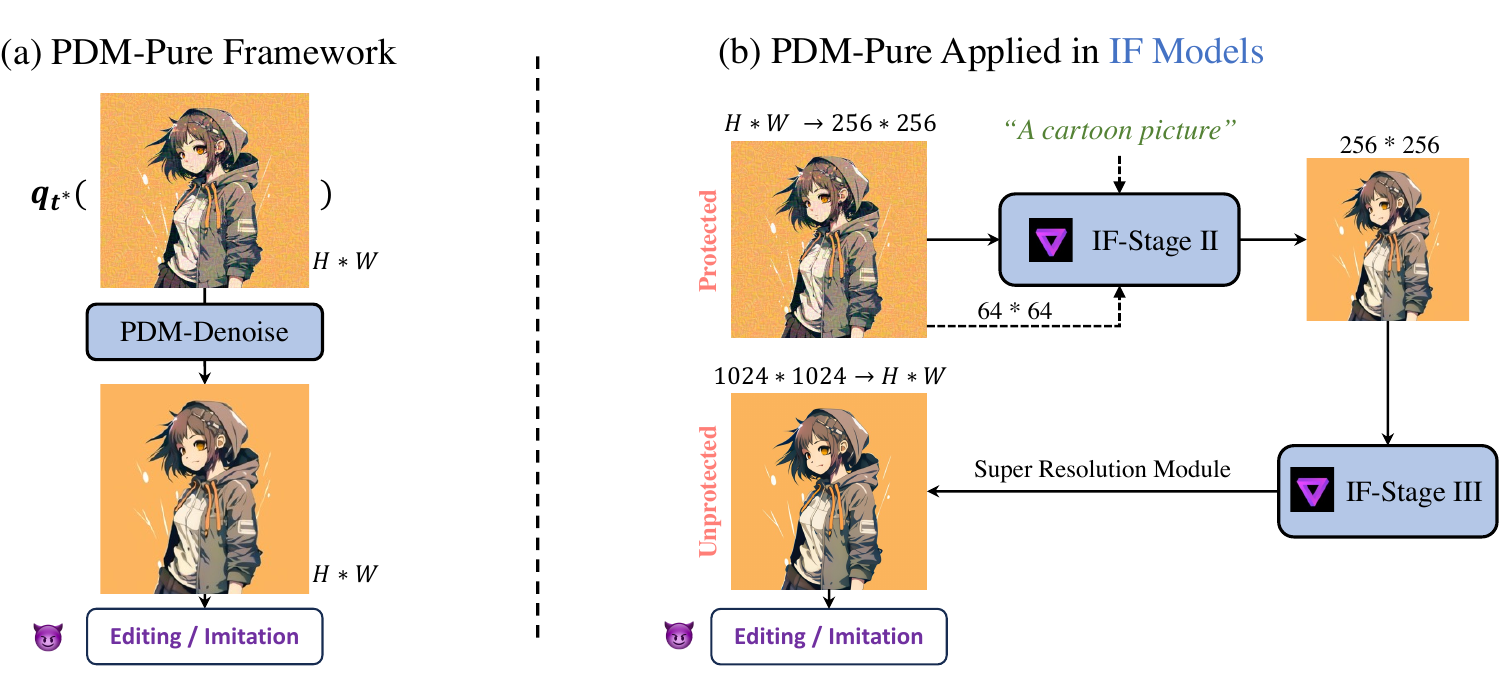}
  \vspace{-5pt}
  \caption{\textbf{PDM-Pure is Easy to Design:} (a) PDM-Pure applies SDEdit~\cite{meng2021sdedit} in the pixel space: it first runs forward diffusion with a small step $t^{*}$ and then runs denoising process. (b) We adapt the framework to DeepFloyd-IF~\cite{deepfloyd}, one of the strongest PDMs. PDM-Pure can effectively remove strong protective perturbations (e.g. $\delta=16/255$). The images we tested are sized $512\times 512$.}

\label{fig:purification_pipeline}
  \vspace{-0.5cm}
\end{figure}

\section{Rethink Adversarial Examples for Diffusion Models}

Adversarial examples of LDMs are widely adopted as a protection mechanism to prevent unauthorized images from being edited or imitated~\cite{glaze, liang2023mist}. However, a significant issue overlooked is that all the adversarial examples in existing work are generated using LDMs, primarily due to the wide impact of the Stable Diffusion; no attempts have been made to attack PDMs. 

This lack of investigation may mislead us to conclude that diffusion models, like most deep neural networks, are vulnerable to adversarial perturbations, and that the algorithms used in LDMs can be transferred to PDMs by simply applying the same adversarial loss in the pixel space formulated as:

\vspace{-0.4cm}
\begin{equation}\label{pixel_diffusion_adversarial_loss}
    \mathcal{L}_{adv}(x) = \mathbb{E}_{t, \epsilon} \mathbb{E}_{x_t \sim q_t(x)}\|\epsilon_{\theta}(x_t, t) -\epsilon \|_2^2
\end{equation}

However, we show through experiments that PDMs are robust against this form of attack (Figure~\ref{fig:attack_various_models}), which means all the existing attacks against diffusion models are, in fact, special cases of attacks against the LDMs only.
Prior to this study, there may have been a prevailing belief that diffusion models could be easily deceived. However, our research reveals an important distinction: it is the LDMs that exhibit vulnerability, while the PDMs demonstrate significantly higher adversarial robustness.
% Before this work, people may think that diffusion models can be easily fooled, but the truth is that only LDMs are, the original PDMs are much more adversarially robust. 
We conduct extensive experiments on popular LDMs and PDMs structures including DiT, Guided Diffusion, Stable Diffusion, and DeepFloyd, and demonstrate in Table~\ref{quant_protect} that only the LDMs can be attacked and PDMs are not that susceptible to adversarial perturbations. More details and analysis can be found in the experiment section.

The vulnerability of the LDMs is caused by the vulnerability of the latent space~\cite{sdsattack}, meaning that although we may set budgets for perturbations in the pixel space, the perturbations in the latent space can be large. In~\cite{sdsattack}, the authors show statistics of perturbations in the latent space over the perturbations in the pixel space and this value $\frac{|\delta_z|}{|\delta_x|}$ can be as large as $10$. In contrast, the PDMs directly work in the pixel space, and thus the injected noise combined with the random Gaussian noise will not easily fool the denoiser as it is trained to be robust to Gaussian noise of different levels.

Almost all the copyright protection perturbations~\cite{glaze, liang2023mist, mist-v2} are based on the insight that it is easy to craft adversarial examples to fool the diffusion models.  We need to rethink the adversarial samples of diffusion models since there are a lot of PDMs that cannot be attacked easily. Next, we show that PDMs can be utilized to purify all adversarial patterns generated by existing methods in Section~\ref{sec:pdm_pure}.  This new landscape poses new challenges to ensure the security and robustness of diffusion-based copyright protection techniques.

\section{PDM-Pure: PDM as a Strong  Universal Purifier}~\label{sec:pdm_pure}
\vspace{-0.4cm}

Given the robustness of PDMs, a natural idea emerges: we can utilize PDMs as a universal purification network. This approach could potentially eliminate any adversarial patterns without knowing the nature of the attacks. We term this framework \textbf{PDM-Pure}, which is a general framework to deal with all the perturbations nowadays. To fully harness the capabilities of PDM-Pure, we need to fulfill two basic requirements: (1) The perturbation shows out-of-distribution pattern as reflected in existing works on adversarial purification/attacks using diffusion models~\cite{nie2022diffusion,diff-pgd} (2) The PDM being used is strong enough to represent $p(x_0)$, which can be largely determined by the dataset they are trained on. 

It is \textbf{effortless} to design a PDM-Pure. The key idea behind this method is to run SDEdit in the pixel space. Given any strong pixel-space diffusion model, we add a small noise to the protected images and run the denoising process (Figure~\ref{fig:purification_pipeline}), and then the adversarial pattern should be removed. The key idea of PDM-Pure is simple. In practice, we need to adjust the pipeline to fit the resolution of the PDMs being used. 

Here, we explain in detail how to adapt DeepFloyd-IF~\cite{deepfloyd}, the strongest open-source PDM as far as we know, for PDM-Pure. DeepFloyd-IF is a cascaded text-to-image diffusion model trained on 1.2B text-image pairs from LAION dataset~\cite{schuhmann2022laion}. It contains three stages named IF-Stage I, II, and III. Here we only use Stage II and III since Stage I works in a resolution of $64$ which is too low. Given a perturbed image $x_{W\times H}$ sized $W\times H$, we first resize it into $x_{64\times 64}$ and $x_{256\times 256}$. Then we use a general prompt  $\mathcal{P}$ to do SDEdit~\cite{meng2021sdedit} using the Stage II model: 
% $x_t = \textbf{IF-II}(x_{t+1}, x_{64\times 64}, p)$

\begin{equation}
    x_t = \textbf{IF-II}(x_{t+1}, x_{64\times 64}, \mathcal{P})
\end{equation}

where $t=T_{\text{edit}}-1, ...,1, 0$, $x_{T_{\text{edit}}}=x_{256\times 256}$. A larger $T_{\text{edit}}$ may be used for larger noise. $x_0$ is the purified image we get in the $256\times 256$ resolution space, where the adversarial patterns should be already purified. We can then use IF Stage III to further up-sample it into $1024\times 1024$ with $x_{1024\times 1024} = \textbf{IF-III}(x_0, p)$. Finally, we can sample into $H\times W$ as we want through downsampling. This whole process is demonstrated in Figure~\ref{fig:purification_pipeline}. After purification, the image is no longer adversarial to the targeted diffusion models and can be effectively used in downstream tasks.

In the main paper, we conduct experiments on purifying protected images sized $512\times 512$. For images with a larger resolution, purifying in the resolution of $256\times 256$ may lose information. In Appendix~\ref{supp:section:pdm_pure_for_higher_resolution} we show PDM-Pure can also applied to purify patches of high-resolution inputs.

\begin{table}[t]
 
  \label{tab:my_label}
 \resizebox{\textwidth}{!}{%
  \centering
  \begin{tabular}{ccccccccc}
\toprule
Methods & AdvDM & AdvDM(-) & SDS(-) & SDS(+) & SDST & Photoguard & Mist & Mist-v2 \\

% \midrule

% $\delta=4/255$ \\
% \midrule

% Crop-Resize & 0 & 0 & 0 & 0 & 0 & 0 & 0 & 0\\

% JPEG & 0 & 0 & 0 & 0 & 0 & 0 & 0 & 0\\

% Adv-Clean & 0 & 0 & 0 & 0 & 0 & 0 & 0 & 0\\

% LDM-Pure& 0 & 0 & 0 & 0 & 0 & 0 & 0 & 0\\

% GrIDPure & 0 & 0 & 0 & 0 & 0 & 0 & 0 & 0\\

% PDM-Pure (ours) & 0 & 0 & 0 & 0 & 0 & 0 & 0 & 0\\

% \midrule
% $\delta=16/255$ \\
\midrule

Before Protection & 166 & 166 & 166 & 166 & 166 & 166 & 166 & 166 \\

After Protection & 297 &221 & 231 & 299 & 322 & 375 & 372 & 370 \\

\midrule

Crop-Resize & 210 & 271 & 228 & 217 & 280 & 295 & 289 & 288\\

JPEG & 296 & 222 & 229 & 297 & 320 & 359 & 351 & 348 \\

Adv-Clean & 243 & 201 & 204 & 244 & 243 & 266 & 282 & 270 \\

LDM-Pure& 300 & 251 & 235 & 300 & 350 & 385 & 380 & 375 \\

GrIDPure & 200 & 182 & 195 & 200 & 210 & 220 & 230 & 210 \\
\rowcolor{LightCyan}
PDM-Pure (ours) & \textbf{161} & \textbf{170} & \textbf{165} & \textbf{159} & \textbf{179} & \textbf{175} & \textbf{178} & \textbf{170}\\

\bottomrule
  \end{tabular}

}
\vspace{10pt}
 \caption{
  \textbf{Quantiative Measurement of Different Purification Methods in Different Scale (FID-score)}: We compute the FID-score of editing purified images over the clean dataset. PDM-Pure is the strongest to remove all the tested protection, under strong protection with $\delta=16$. GrIDPure~\cite{zhao2023can} can also do reasonable protection, but the performance is limited because the PDM they used is not strong enough.
  }
  \label{quant_purify}
  \vspace{-0.8cm}
\end{table}

\section{Experiments}

In this section, we conduct experiments on various attacking methods and various models to support the following two conclusions:

\begin{itemize}[parsep=0pt,topsep=0pt,leftmargin=12pt]
    \item \textbf{(C1)}: PDMs are much more adversarial robust than LDMs, and PDMs can not be effectively attacked using all the existing attacks for LDMs.
    \item \textbf{(C2)}: PDMs can be applied to effectively purify all of the existing protective perturbations. Our PDM-Pure based on DeepFloyd-IF shows state-of-the-art purification power.
    % \item \textbf{(C3)}: Pixel is a barrier for us to achieve real protection against diffusion-based mimicry. PDM-Pure can make the protective perturbation no more protective, and there is currently no effective way to attack PDMs.
\end{itemize}

\subsection{Models, Datasets, and Metrics} 
The models we used can be categorized into LDMs and PDMs. For LDMs, we use Stable Diffusion V-1.4, V-1.5 (SD-V-1.4, SD-V-1.5)~\cite{ldm}, and Diffusion Transformer (DiT-XL/2)~\cite{dit}, and for PDMs we use Guided Diffusion (GD)~\cite{guideddiffusion} trained on ImageNet~\cite{deng2009imagenet}, and DeepFloyd Stage I and Stage II~\cite{deepfloyd}. 

For models trained on the ImageNet (DiT, GD), we run adversarial attacks and purification on a 1k subset of the ImageNet validation dataset. For models trained on LAION, we run tests on the dataset proposed in~\cite{sdsattack}, which includes $400$ cartoon, artwork, landscape, and portrait images. The metrics for testing the quality of generated images are included in the Appendix.

For protection methods, we consider almost all the representative approaches, including AdvDM~\cite{advdm}, SDS~\cite{sdsattack}, Mist~\cite{liang2023mist}, Mist-v2~\cite{mist-v2}, Photoguard~\cite{salman2023raising} and Glaze~\cite{glaze}. We also test the methods in the design space proposed in ~\cite{sdsattack}, including  SDS(-), AdvDM(-), and SDST. In contrast to other existing methods, they are based on gradient descent and have shown great performance in deceiving the LDMs.

\subsection{(C1) PDMs are Much More Robust Than We Think} 

In Table~\ref{quant_protect}, we attack different LDMs and PDMs with one of the most popular adversarial loss~\cite{mist-v2} in Equation~\ref{semantic_loss} and Equation~\ref{pixel_diffusion_adversarial_loss}, which can be interpreted as fooling the denoiser using a Monte-Carlo-based loss. Given the attacked samples, we test the SDEdit results on the attacked samples, which can be generally used to test whether the samples are adversarial for the diffusion model or not. We use FID-score~\cite{fid}, SSIM~\cite{ssim}, LPIPS~\cite{lpips}, and IA-Score~\cite{la-score} to measure the quality of the attack. If the quality of generated images decreases a lot compared with editing the clean images, then the attack is successful. We can see that LDMs can be easily attacked, while PDMs are quite robust; the quality of the edited images is still good. We also show some visualizations in Figure~\ref{fig:attack_various_models}, which illustrates that the perturbation will affect the LDMs but not the PDMs.

To further investigate how robust PDM is, we test other advanced attacking methods, including the End-to-End Diffusion Attacks (E2E-Photoguard) proposed in~\cite{salman2023raising} and the Improved Targeted Attack (ITA) proposed in ~\cite{mist-v2}. Though the End-to-End attack is usually impractical to run, it shows the strongest performance to attack LDMs.  We find that both attacks are not successful in PDM settings. We show attacked samples and edited samples in Figure~\ref{fig:attack_various_models} as well as the Appendix. In conclusion, existing adversarial attack methods for diffusion models can only work for the LDMs, and PDMs are more robust than we think.

\subsection{(C2) PDM-Pure: A Universal Purifier that is Simple yet Effective}

PDM-Pure is simple: basically, we just run SDEdit to purify the protected image in the pixel space. Given our assumption that PDMs are quite robust, we can use PDMs trained on large-scale datasets as a universal black-box purifier. We follow the model pipeline introduced in Section~\ref{sec:pdm_pure} and purify images protected by various methods in Table~\ref{quant_purify}.

\begin{figure}[H]
  \centering
\includegraphics[width=0.99\linewidth]{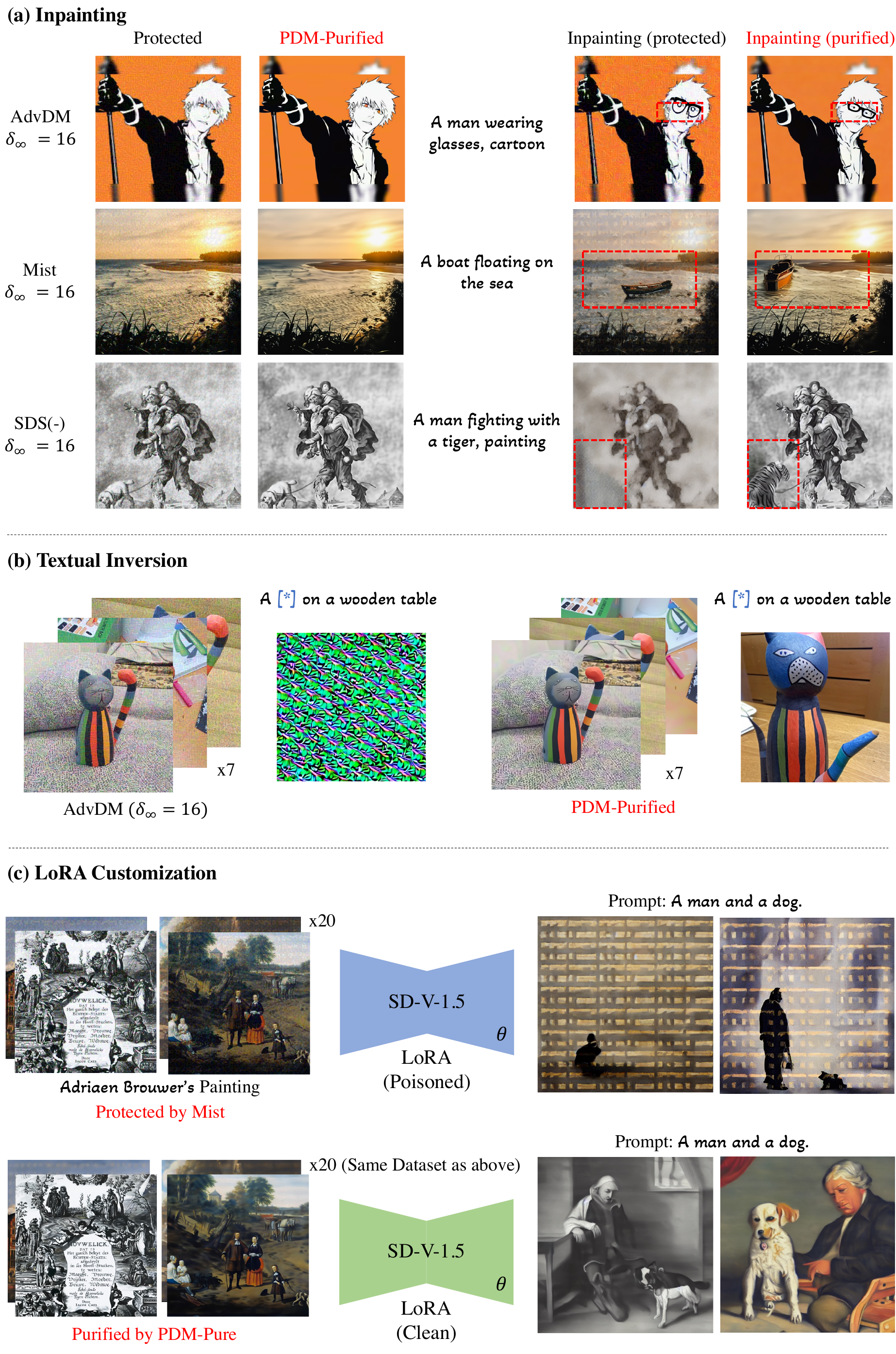}
  \vspace{-5pt}
  \caption{
  \textbf{PDM-Pure makes the Protected Images no more Protected:} Here we show qualitative results of PDM-Pure on three scenarios where unauthorized editing may occur: (a) Inpainting, (b) Text-Inversion~\cite{textualinversion} and (c) LoRA customization~\cite{lora}. While the protected images incur bad generation quality, the purified ones can fully bypass the protection.
  }
  % \caption{\textbf{Qualitative Results of PDM-Pure in Inpainting:} PDM-Pure can effectively remove the adversarial patterns in various protection methods, here we show image inpainting results on three typical protection methods: AdvDM~\cite{advdm}, Mist~\cite{liang2023mist} and SDS(-)~\cite{sdsattack} on Stable Diffusion V-1.5. We show results on quite strong attacks with budget $\delta_{\infty}=16/255$. We can see the figures are no more adversarial after PDM-Pure, resulting in much better inpainting results. (inpainting regions are indicated by the the bounding boxes; zoom in on screen for better observation)}

  \label{fig:purification_results}
\end{figure}

PDM-Pure is effective: from Table~\ref{quant_purify} we can see that the purification will remove adversarial patterns for all the protection methods we tested, largely decreasing the FID score for the SDEdit task. Also, we test the protected images and purified images in more tasks including Image Inpainting~\cite{song2020score}, Textual-Inversion~\cite{textualinversion}, and LoRA customization~\cite{lora} in Figure~\ref{fig:purification_results}. Both qualitative and quantitative results show that the purified images are no more adversarial and can be effectively edited or imitated in different tasks without any obstruction. 

Also, PDM-Pure shows SOTA results compared with previous purification methods, including some simple purifiers based on compression and filtering like Adv-Clean, crop-and-resize, JPEG Compression, and  SDEdit-based methods like GrIDPure~\cite{zhao2023can}, which uses patchified SDEdit with a GD~\cite{guideddiffusion}. We also add LDM-Pure as a baseline to show that LDMs can not be used to purify the protected images. For GrIDPure, we use Guided-Diffusion trained on ImageNet to run patchified purification. All the experiments are conducted on the datasets collected in ~\cite{sdsattack} under the resolution of $512\times 512$. Results for higher resolutions are presented in Appendix~\ref{supp:section:pdm_pure_for_higher_resolution}.

% \subsection{(C3) Pixel is A Barrier: What Should We Do in the Future?}

% \chen{why is this in the experiment section? Are any experiments involved?} Pixel is a barrier for us to do real protection against adversarial attacks. Since PDMs are quite robust, they cannot be easily attacked and can even be used to purify the protective perturbations, challenging the current assumption for safety protection of generative diffusion models. The community should rethink the problem of adversarial samples for generative diffusion models and rethink can we rely on them to protect unauthorized images. Hence, diffusion models turn out to be quite robust, more research should be conducted to study them and the reason behind them. If the robustness can be verified and guaranteed, we may rely on it as a new structure for many other tasks.\haotian{move it the the conclusion}

\section{Conclusions and Future Directions}

In this paper, we present novel insights that while many studies demonstrate the ease of finding adversarial samples for Latent Diffusion Models (LDMs), Pixel Diffusion Models (PDMs) exhibit far greater adversarial robustness than previously assumed. We are the first to investigate the adversarial samples for PDMs, revealing a surprising discovery that existing attacks fail to fool PDMs. Leveraging this insight, we propose utilizing strong PDMs as universal purifiers, resulting in PDM-Pure, a simple yet effective framework that can generate protective perturbations in a black-box manner.

Pixel is a barrier for us to do
real protection against adversarial attacks. Since PDMs are quite robust, they cannot be easily attacked.
PDMs can even be used to purify the protective perturbations, challenging the current assumption for
the safe protection of generative diffusion models. We advocate rethinking the problem of
adversarial samples for generative diffusion models and 
unauthorized image protection based on it. 
% Diffusion models turn out to be quite robust, 
More rigorous study can be
conducted to better understand the mechanism behind the robustness of PDMs. Furthermore, we can utilize it as a new structure for many other tasks

%% test citation
% \citep{diff-pgd, sdsattack}

{
\small
\bibliographystyle{abbrvnat}
\bibliography{bibliography}
}

\newpage  
\tableofcontents

\appendix
\newpage
\part*{\Huge Appendix}

% \section{Possible Questions}

% \chen{do we really want to include this section?}
% \paragraph{Q:}\textit{What is the Difference between PDM-Pure and Diff-Pure~\cite{nie2022diffusion}?}

% Diff-Pure~\cite{nie2022diffusion} is proposed to purify 

% \paragraph{Q:}\textit{Can Adaptive Attacks be Used to Attack PDM-Pure?}

% Since PDM-Pure can be used to purify protective perturbations, can we develop stronger adaptive attacks targeting the purification process? Actually, ~\cite{zhao2023can} already did experiments to adaptively attack DiffPure and the answer is that the protection still does not work. It further reflects our conclusion that PDM is robust.

\section{Details about Different Diffusion Models in this Paper}

Here we introduce the diffusion models used in this work, which cover different types of diffusion (LDM, PDM), different training datasets, different resolutions, and different model structures (U-Net, Transformer):

\paragraph{Guided Diffusion (PDM)} We use the implementation and checkpoint from \url{https://github.com/openai/guided-diffusion}, the Guided Diffusion models we used are trained on ImageNet~\cite{deng2009imagenet} in resolution $256\times 256$, the editing results are tested on sub-dataset of ImageNet validation set sized 500.

\paragraph{IF-Stage I (PDM)} This is the first stage of the cascaded DeepFloyd IF model~\cite{deepfloyd} from \url{https://github.com/deep-floyd/IF}. It is trained on LAION 1.2B with text annotation. It has a resolution of $64\times 64$. the editing results are tested on the image dataset introduced in ~\cite{sdsattack}, including 400 anime, portrait, landscape, and artwork images.

\paragraph{IF-Stage II (PDM)} This is the second stage of the cascaded DeepFloyd IF model~\cite{deepfloyd} from \url{https://github.com/deep-floyd/IF}. It is a conditional diffusion model in the pixel space with $256\times 256$, which is conditioned on $64\times 64$ low-resolution images. During the attack, we freeze the image condition and only attack the target image to be edited.

% \chen{how about IF-Stage III?}\haotian{stage iii is for super-resolution in 1024*1024, the pixel-space version is not opensource, the opensource one uses SD}

\paragraph{Stable Diffusion V-1.4 (LDM)}
It is one of the most popular LDMs from \url{https://huggingface.co/CompVis/stable-diffusion-v1-4}, also trained on text-image pairs, which has been widely studied in this field. It supports resolutions of $256\times 256$ and $512\times 512$, both can be easily attacked. The encoder first encodes the image sized $H\times W$ into the latent space sized $4\times H/4 \times W/4$, and then uses U-Net combined with cross-attention to run the denoising process.

\paragraph{Stable Diffusion V-1.5 (LDM)}
It has the same structure as Stable Diffusion V-1.4, which is also stronger since it is trained with more steps, from \url{https://huggingface.co/runwayml/stable-diffusion-v1-5}.

\paragraph{DiT-XL (LDM)} It is another popular latent diffusion model, that uses the backbone of the Transformer instead of the U-Net. We use the implementation from the original repository \url{https://github.com/facebookresearch/DiT/}.

\section{Details about Different Protection Methods in this Paper}
We introduce different protection methods tested in this paper, of which all the original versions are designed for LDMs. All the adversarial attacks work under the white box settings of PGD-attack, varying from each other with different adversarial losses:

\paragraph{AdvDM} AdvDM is one of the first adversarial attacks proposed in ~\cite{advdm}, it used a Monte-Carlo-based adversarial loss which can effectively attack the latent diffusion models, we also call this loss semantic loss:

\begin{equation}\label{appendix:semantic_loss}
        \mathcal{L}_{S}(x) = \mathbb{E}_{t, \epsilon} \mathbb{E}_{z_t \sim q_t(\mathcal{E}_{\phi}(x))}\|\epsilon_{\theta}(z_t, t) -\epsilon \|_2^2
\end{equation}

\paragraph{PhotoGuard} PhotoGuard is proposed in ~\cite{salman2023raising}, it takes the encoder, making the encoded image close to a target image $y$, we also call it textural loss:

\begin{equation}\label{appendix:tex-loss}
        \mathcal{L}_{T}(x) = -\|\mathcal{E}_{\phi}(x) -\mathcal{E}_{\phi}(y) \|_2^2
    \end{equation}

\paragraph{Mist} Mist~\cite{liang2023mist} finds that ${L}_{T}(x)$ can better enhance the attacks if the target image $y$ is chosen to be periodical patterns, the final loss combined ${L}_{T}(x)$ and ${L}_{S}(x)$:

\begin{equation}
    \mathcal{L} = \lambda {L}_{T}(x) + {L}_{S}(x)
\end{equation}

\paragraph{SDS(+)} Proposed in ~\cite{sdsattack}, it is proven to be a more effective attack compared with the original AdvDM, where the gradient $\nabla_x\mathcal{L}(x)$ is expensive to compute. By using the score distillation-based loss, it shows good performance and remains effective at the same time:

\begin{equation}\label{sds_equation}
    \nabla_{x}\mathcal{L}_{SDS}(x) =  \mathbb{E}_{t, \epsilon}\mathbb{E}_{z_t} \left[\lambda(t) (\epsilon_{\theta}(z_t, t) -\epsilon)\frac{\partial z_t}{\partial x_t}\right]
\end{equation}

\paragraph{SDS(-)} Similar to SDS(+), it swaps gradient ascent in the original PGD with gradient descent, which turns out to be even more effective.

\begin{equation}\label{sds_equation}
    \nabla_{x}\mathcal{L}_{SDS(-)}(x) = -\mathbb{E}_{t, \epsilon}\mathbb{E}_{z_t} \left[\lambda(t) (\epsilon_{\theta}(z_t, t) -\epsilon)\frac{\partial z_t}{\partial x_t}\right]
\end{equation}

\paragraph{Mist-v2} It was proposed in ~\cite{mist-v2} using the Improved Targeted Attack (ITA), which turns out to be very effective, especially when the limit budget is small. It is also more effective to attack LoRA:

\begin{equation}
     \mathcal{L}_{S}(x) = \mathbb{E}_{t, \epsilon} \mathbb{E}_{z_t \sim q_t(\mathcal{E}_{\phi}(x))}\|\epsilon_{\theta}(z_t, t) -z_0 \|_2^2
\end{equation}

where $z_0 = \mathcal{E}(y)$ is the latent of a target image, which is the same as the typical image used in Mist.

\paragraph{Glaze} It is the most popular protection claimed to safeguard artists from unauthorized imitation~
\cite{glaze} and is widely used by the community. while it is not open-sourced, it also attacks the encoder like the Photoguard. Here we only test it in the purification stage, where we show that the protection can also be bypassed.

% \paragraph{MetaCloak} It is proposed in~\cite{metacloak} which is strong against different kinds of purification, we also test it in the PDM-Pure stage.

\paragraph{End-to-End Attack} It is also first proposed in ~\cite{salman2023raising}, which attacks the editing pipeline in a end-to-end manner. Although it is strong, it is not practical to use and does not show dominant privilege compared with other protection methods.

\section{Details about The Evaluation Metrics}\label{supp:section:eval_metrics}

 Here we introduce the quantitative measurement we used in our experiments: 

 \begin{itemize}
     \item We measure the SDEdit results after the adversarial attacks using Fréchet Inception Distance (FID)~\citep{fid} over the relevant datasets (for model trained on ImageNet such as GD~\cite{guideddiffusion} and DiT~\cite{dit} we use a sub-dataset of ImageNet as the relevant dataset, for those trained on LAION, we use the collected dataset to calculate the FID). We also use Image-Alignment Score (IA-score)~\citep{la-score}, which can be used to calculate the cosine-similarity between the CLIP embedding of the edited image and the original image. Also, we use some basic evaluations, where we calculate the Structural Similarity (SSIM)~\citep{ssim} and Perceptual Similarity (LPIPS)~\citep{lpips} compared with the original images.

     \item To measure the purification results, we test the Fréchet Inception Distance (FID)~\citep{fid} over the collected dataset compared with the dataset generated by running SDEdit over the purified images in the strength of $0.3$.
 \end{itemize}
 
 % (2) To measure the protection results, we use FID, LPIPS, Peak Signal-to-Noise Ratio (PSNR)~\citep{psnr}, and Image-Alignment Score (IA-score)~\citep{la-score} which calculated the cosine-similarity between the CLIP embedding of the protected image and the original image. Also, we have human evaluations which are collected using surveys, which is a more convincing way to evaluate the quality of protections, more settings can be found in the appendix.

\section{Details about Different Purification Methods}\label{supp:section:purification_baselines}

\paragraph{Adv-Clean:} \url{https://github.com/lllyasviel/AdverseCleaner}, a training-free filter-based method that can remove adversarial noise for a diffusion model, it works well to remove high-frequency noise.

\paragraph{Crop $\&$ Resize:} we first crop the image by $20\%$ and then resize the image to the original size, it turns out to be one of the most effective defense methods \citep{liang2023mist}.

\paragraph{JPEG compression:} \citep{sandoval2023jpeg} reveals that JPEG compression can be a good purification method, and we adopt the $65\%$ as the quality of compression in \citep{sandoval2023jpeg}.

\paragraph{LDM-Pure:} We also try to use LDMs to run SDEdit as a naive purifier, sadly it cannot work, because the adversarial protection transfers well between different LDMs.

\paragraph{GrIDPure:} It is proposed in ~\cite{zhao2023can} as a purifier, GrIDPure first divides an image into patches sized $128\times 128$, and then purifies the $9$ patches sized $256\times 256$. Also, it combined the four corners sized $128\times 128$ to purify it so we have $10$ patches to purify in total. After running SDEdit with a small noise (set to $0.1 T$), we reassemble the patches into the original size, pixel values are assigned using the average values of the patches they belong to. More details can be seen in ~\cite{zhao2023can}.

\section{More Experimental Results}

In this section, we present more experimental results.

\subsection{More Visualizations of Attacking PDMs}

\begin{figure}
    \centering
    \includegraphics[width=.99\textwidth]{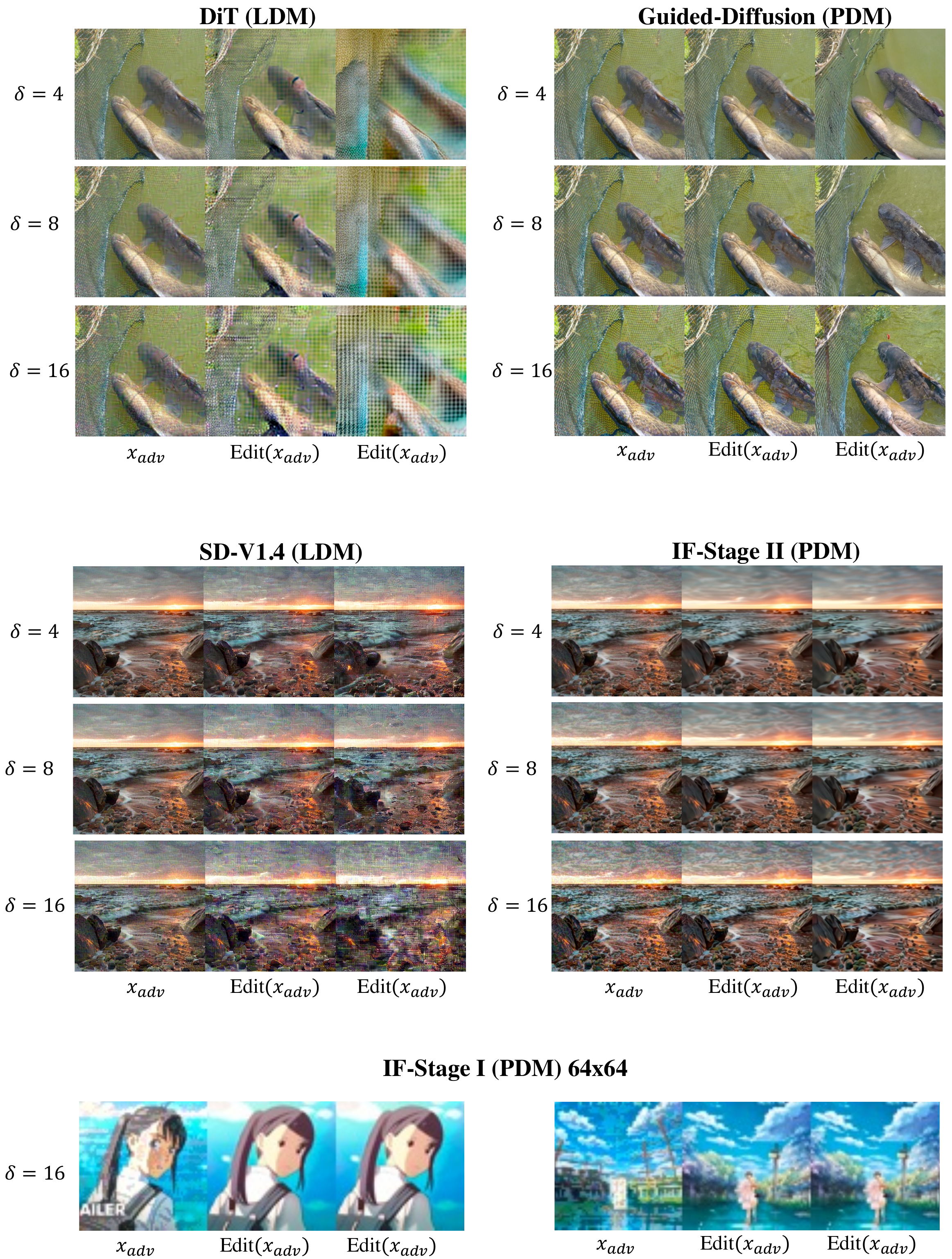}
    \caption{\textbf{PDMs cannot be Attacked as LDMs}: we conduct experiments on various models with various budgets, even the largest budget will not affect the PDMs, showing that PDMs are adversarially robust. For each block, the first column is the attacked image, and the second and third columns are edited images, where the third column adopts larger editing strength.}
    \label{fig:supp:attacking_pdms}
\end{figure}

We show more results of attacking LDMs and PDMs in Figure~\ref{fig:supp:attacking_pdms}, where we attack them with different budget $\delta=4,8,16$. We can see all the LDMs can be easily attacked, while PDMs cannot be attacked, even the largest perturbations will not fool the editing process. Actually, the editing process is trying to purify the strange perturbations.

\subsection{More Visualizaitons of PDM-Pure and Baseline Methods}

\begin{figure}
    \centering
    \includegraphics[width=.99\textwidth]{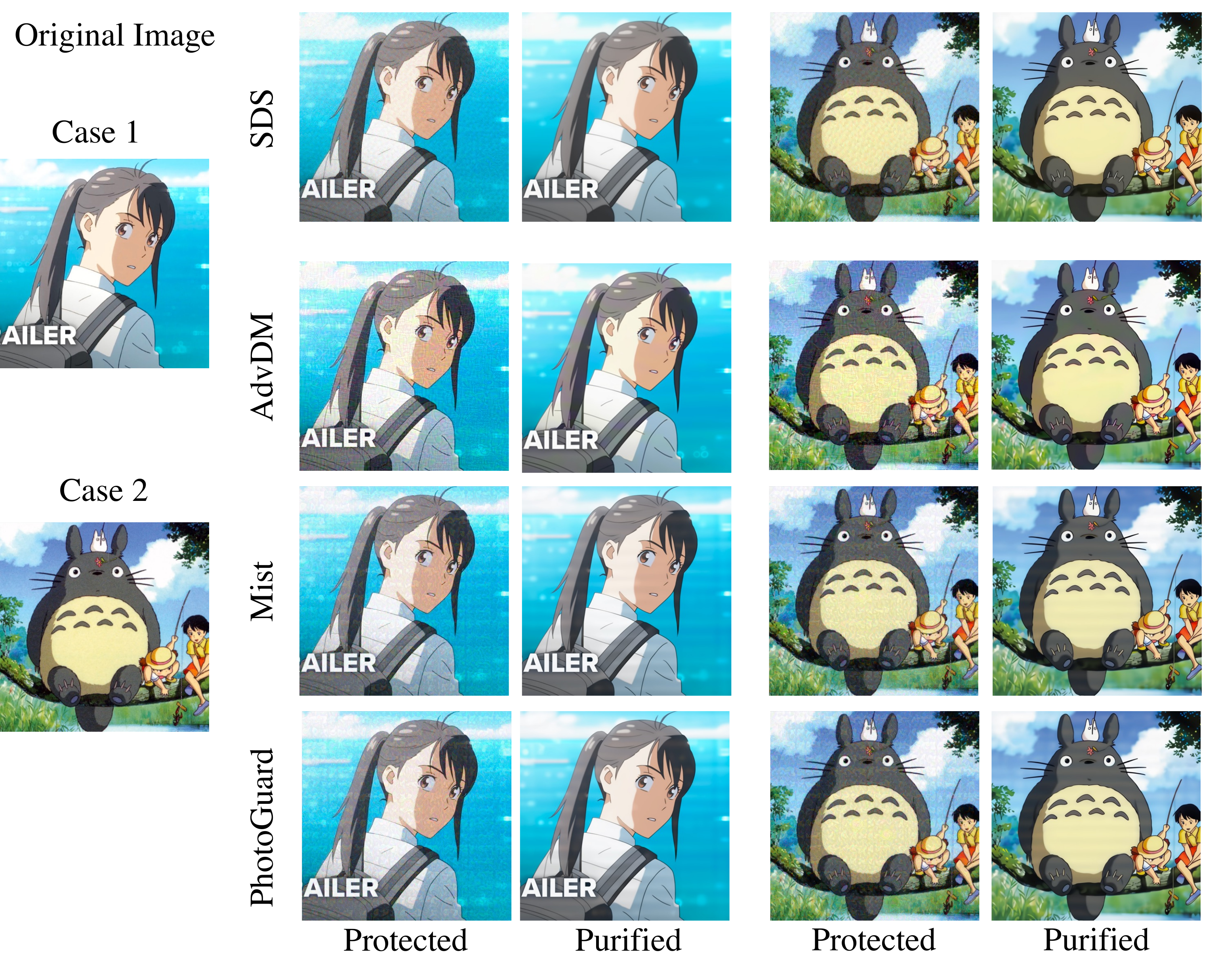}
    \caption{\textbf{More Purification Results of PDM-Pure}: we show purification results compared with the clean image, working on SDS, AdvDM, Mist, and PhotoGuard.}
    \label{fig:supp:pdm_pure_visualize}
\end{figure}

We show more qualitative results of the proposed PDM-Pure based on IF. First, we show purified samples of PDM-Pure in Figure.~\ref{fig:supp:pdm_pure_visualize}, from which we can see that PDM-Pure can remove large protective perturbations and largely preserve details. 

\begin{figure}
    \centering
\includegraphics[width=.99\textwidth]{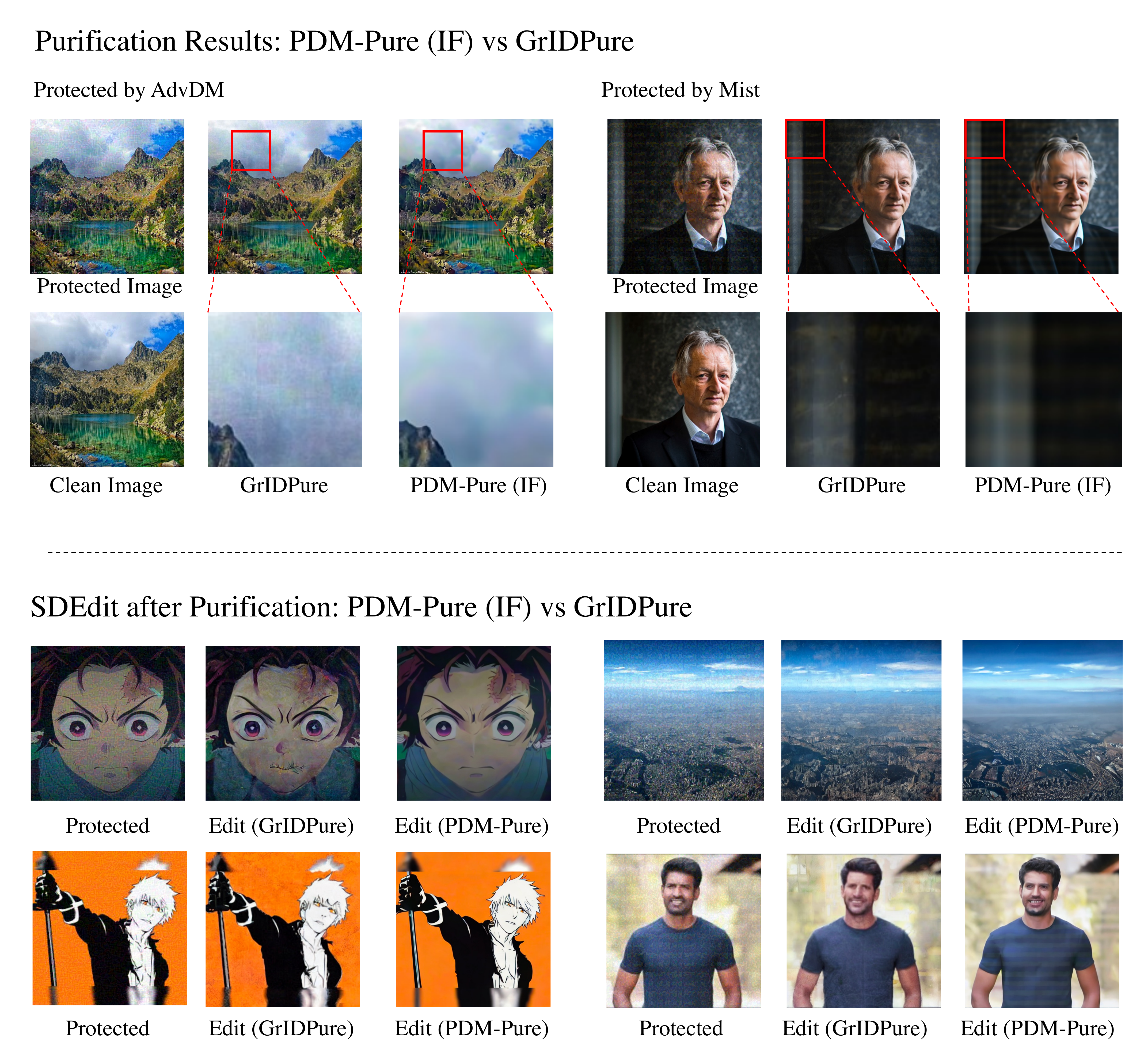}
    \caption{
    % \chen{why is the protected one so good?}\haotian{oh, it is not editing results, it is just the original image} 
    \textbf{PDM-Pure vs GrIDPure}: PDM-Pure is better than GrIDPure, especially when the adversarial pattern is strong such as AdvDM. The bottom half of this figure shows the editing results of purified images, we can see that the editing results of GrIDPure still show somewhat artifacts.}
    \label{fig:supp:pdm_pure_compared with gridpure}
\end{figure}

Compared with GrIDPure~\cite{zhao2023can}, we find that PDM-Pure shows better results when the noise is large and colorful, as is illustrated in Figure~\ref{fig:supp:pdm_pure_compared with gridpure}. Also, though GrIDPure merges patches, it still shows boundary lines between patches. 

\begin{figure}
    \centering
    \includegraphics[width=.99\textwidth]{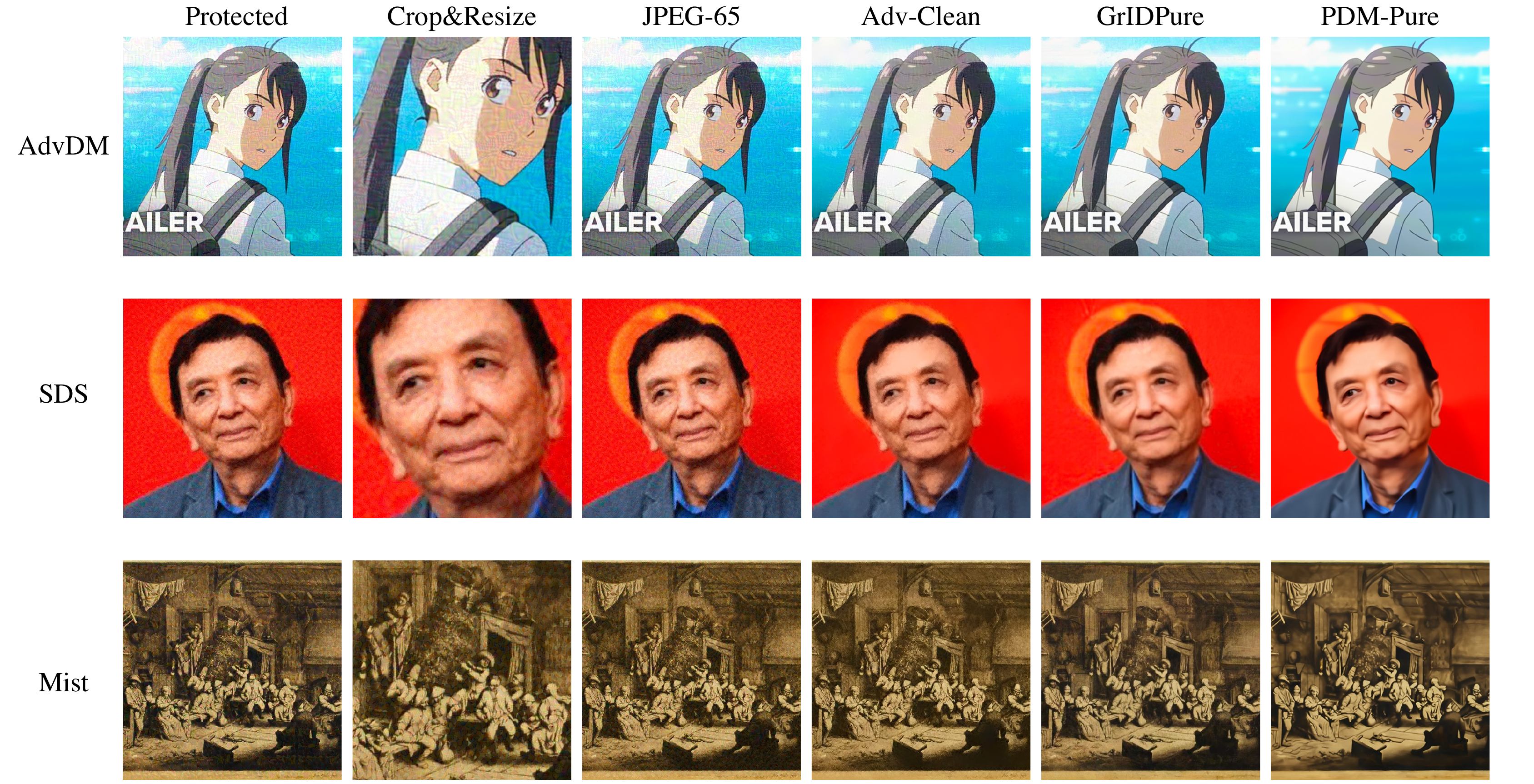}
    \caption{
    % \chen{all the methods seem to be effective}\haotian{baselines still show adversarial patterns}
    \textbf{PDM-Pure Compared With Other Baseline Methods}: we test all the baselines on three typical kinds of protection methods, with $\delta=16/255$. PDM-Pure shows strong performance.}
    \label{fig:supp:different_purification_methods}
\end{figure}

Compared with other baseline purification methods such as Adv-Clean, Crop-and-Resize, and JPEG compression, PDM-Pure shows much better results (Figure~\ref{fig:supp:different_purification_methods}) for different kinds of protective noise, showing that it is capable to serve as a universal purifier. We choose AdvDM, Mist, and SDS as the representative of three kinds of protection. 

\subsection{More Visualizaitons of PDM-Pure for Downstreaming Tasks}

After applying PDM-Pure to the protected images, they are no longer adversarial to LDMs and can be easily edited or imitated. Here we will demonstrate more results on editing the purified images on downstream tasks.

\begin{figure}
    \centering
    \includegraphics[width=.99\textwidth]{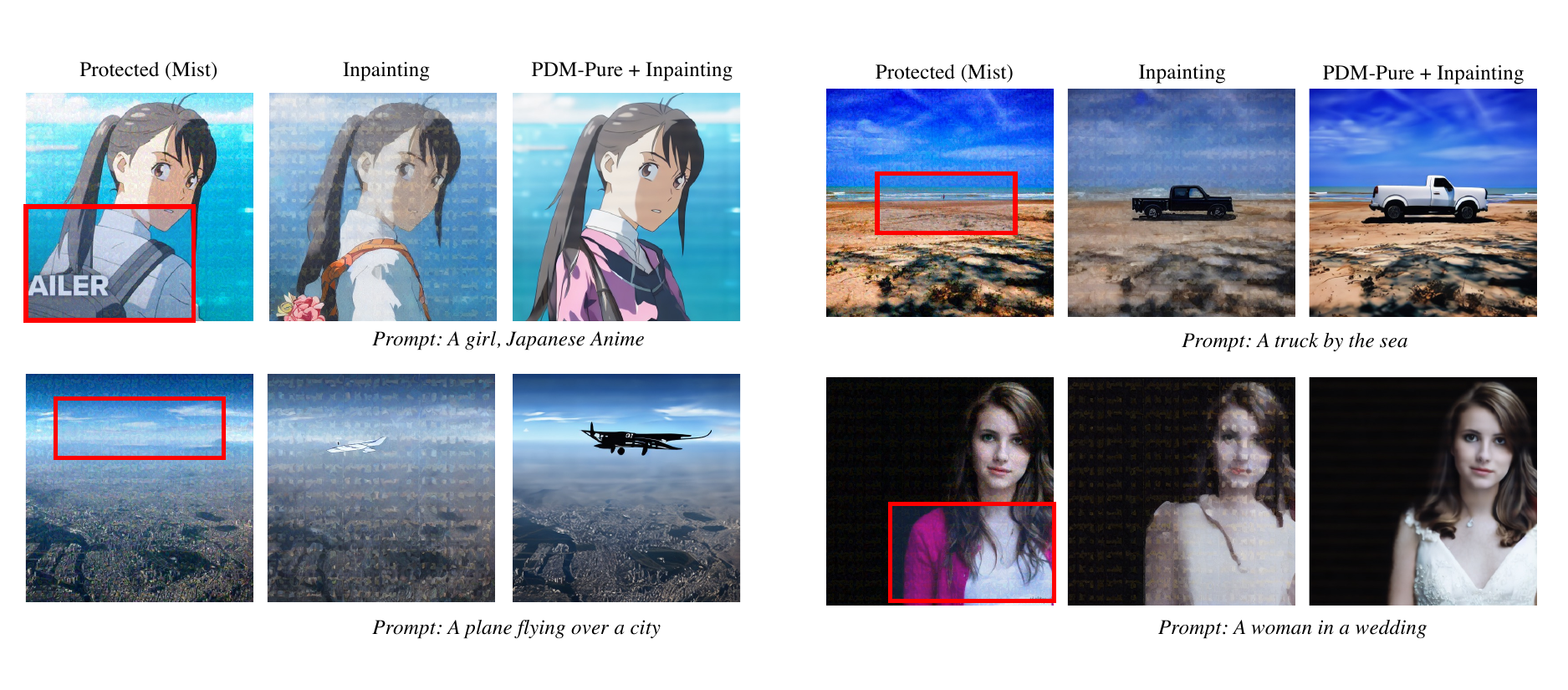}
    \caption{\textbf{More Results of PDM-Pure Bypassing Protection for Inpainting}: after purification, the protected images can be easily inpainted with a high quality. The protective perturbations are generated using Mist with $\delta=16/255$, which is a strong perturbation.}
    \label{fig:supp:pdm_pure_inpainting}
\end{figure}

In Figure~\ref{fig:supp:pdm_pure_inpainting}, we show more results to prove that the purified images can be edited easily, and the quality of editing results is high. It means that PDM-Pure can bypass the protection very well for inpainting tasks.

\begin{figure}
    \centering
    \includegraphics[width=.99\textwidth]{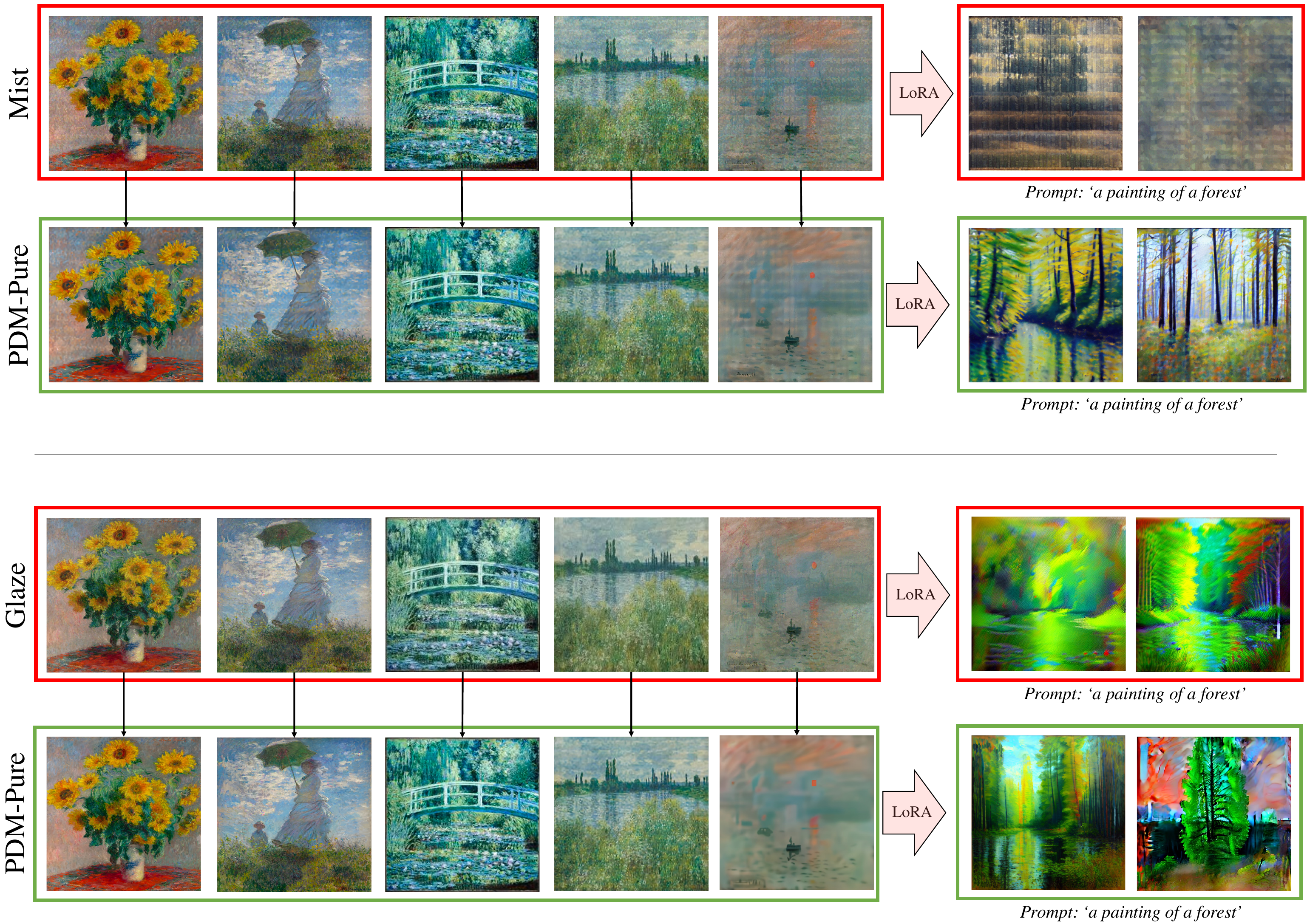}
    \caption{\textbf{More Results of PDM-Pure Bypassing Protection for LoRA}: after purification, the protected images can be imitated again. Here we show examples using $5$ paintings of Claude Monet.}
    \label{fig:supp:pdm_pure_lora}
\end{figure}

In Figure~\ref{fig:supp:pdm_pure_lora} we show more results on purifying Mist~\cite{liang2023mist} and Glaze~\cite{glaze} perturbations, and then running LoRA customized generation. From the figure, we can see that PDM-Pure can make the protected images easy to imitate again.

\section{PDM-Pure For Higher Resolution}\label{supp:section:pdm_pure_for_higher_resolution}

In this paper, we mainly apply PDM-Pure for images sized $512\times 512$, which is also the most widely used resolution for latent diffusion models. When the resolution is $512\times 512$, running SDEdit using Stage II of DeepFloyd makes sense, while if the image size becomes larger, details may be lost because of the downsampling. Hopefully, we can still do purification patch-by-patch with PDM-Pure, in Figure~\ref{supp:section:pdm_pure_for_higher_resolution} we show purification results on images with different resolutions protected by Glaze~\cite{glaze}.

\begin{figure}
    \centering
    \includegraphics[width=.99\textwidth]{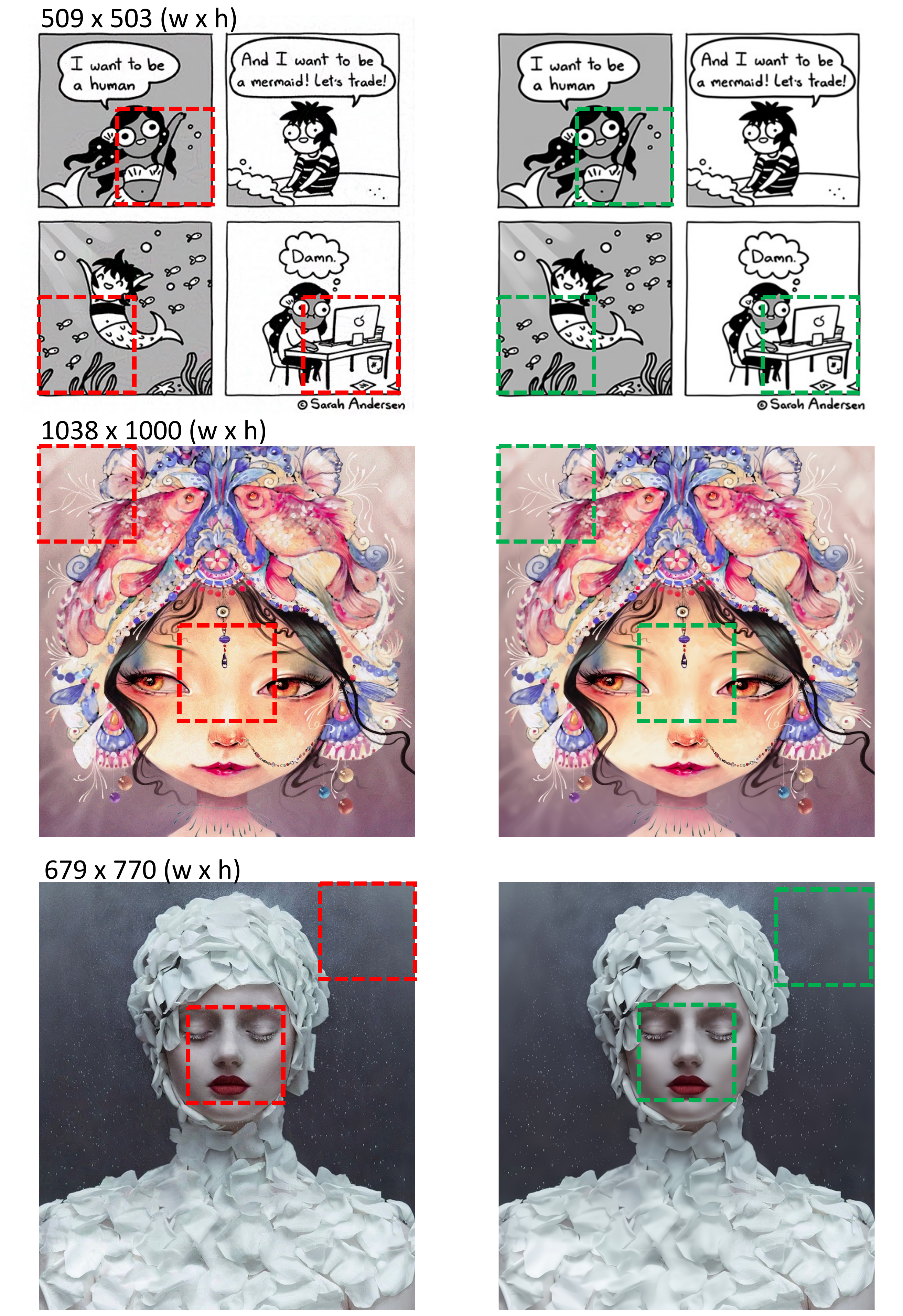}
    \caption{\textbf{PDM-Pure Working On Images with Higher Resolution}: we show the results of applying PDM-Pure for images with higher resolutions, the images are protected using Glaze~\cite{glaze}. We can see from the figure that the adversarial patterns (in red box) can be effectively purified (in green box). Zoom in on the computer for a better view.}
    \label{fig:supp:pdm_pure_larger_image}
\end{figure}

\section{Ablations of $t^*$ in PDM-Pure}

The PDM-Pure on DeepFloyd-IF we used in this paper uses the default settings of SDEdit with $t^* = 0.1 T$. And we respace the diffusion model into $100$ steps, so we only need to run $10$ denoising steps. It can be run on one A6000 GPU, occupying $~22G$ VRAM in $30$ seconds.

\begin{figure}
    \centering
    \includegraphics[width=.99\textwidth]{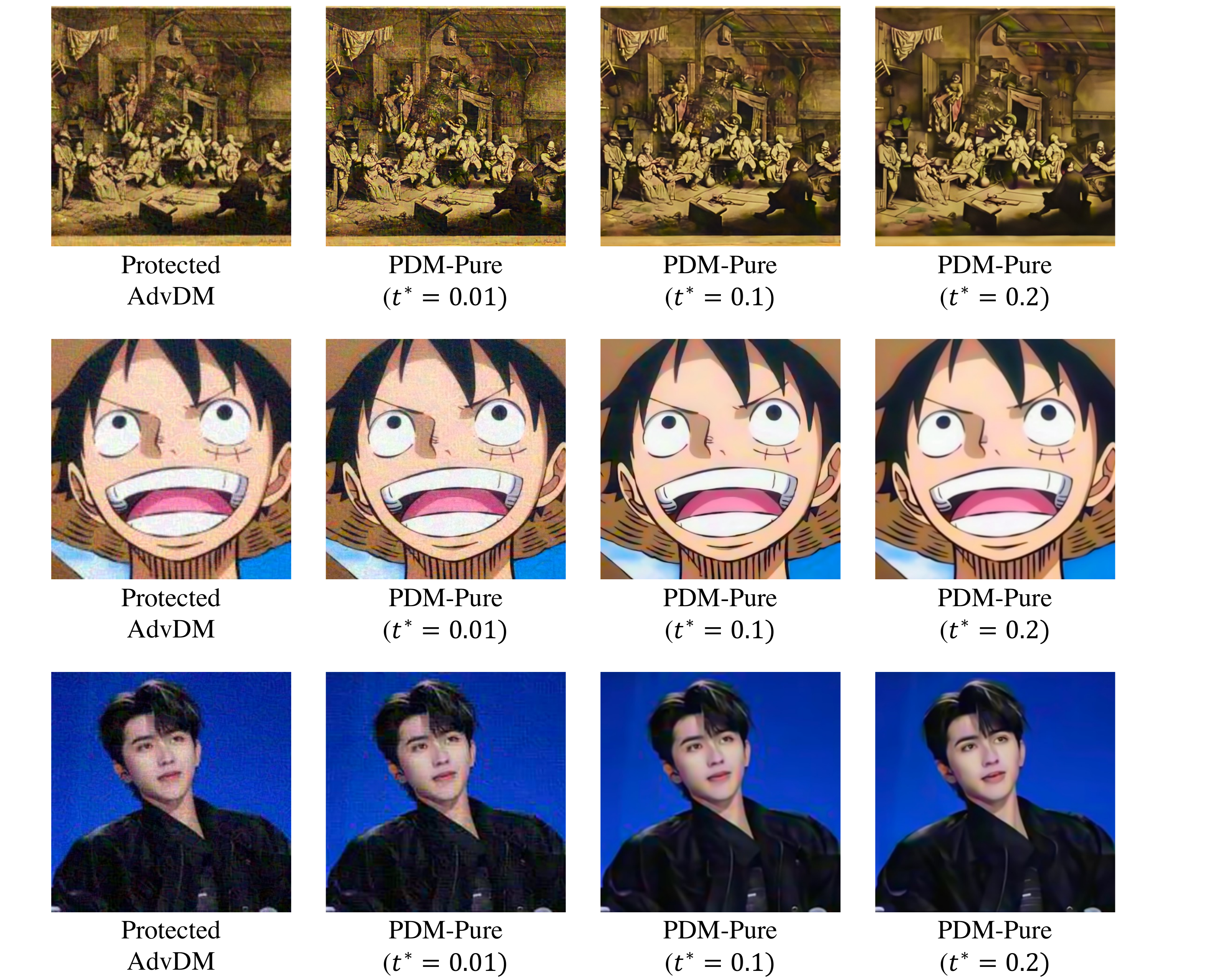}
    \caption{\textbf{PDM-Pure with Different $t^*$}}
    \label{fig:supp:pdm-pure_ablation}
\end{figure}

Here we show some ablation about the choice of $t^*$. In fact, in many SDEdit papers, $t^*$ can be roughly defined by trying, different $t^*$ that can be used to purify different levels of noise. We try $t^*=0.01, 0.1, 0.2$, in Figure~\ref{fig:supp:pdm-pure_ablation} we can see that when $t^*=0.01$ the noise is not fully purified, and when $t^*=0.2$, the details in the painting are blurred. It should be noted that the sweet point for different images and different noises can be slightly different, so it will be more useful to do some trials before purification.

% \begin{table}[]
%     \centering
%     \begin{tabular}{cccccc}
%     \toprule
%         Model Name & Type & Structure & Resolution & Dataset & Resolution \\
%     \midrule
%       SD-V1.4~\cite{ldm}   & LDM & U-Net & 512 & LION & \\
%       SD-V1.5~\cite{ldm}  & LDM &U-Net & 512 & LION & \\
%       % SD-V2.1~\cite{ldm}  & LDM &U-Net & 512 & LION & \\
%       SD-XL~\cite{sdxl}  & LDM & U-Net & 512 & LION & \\
%       DeepFloyd Stage-I~\cite{deepfloyd} & PDM & U-Net & 64 & LION & \\
%       DeepFloyd Stage-II~\cite{deepfloyd} & PDM(c) & U-Net & 256 & LION & \\
%       Guided Diffusion a~\cite{guideddiffusion} & PDM & U-Net & 256 & ImageNet & \\
%       Guided Diffusion b~\cite{guideddiffusion}  & PDM & U-Net & 256 & LSUN-Bedroom & \\
%       Guided Diffusion c~\cite{guideddiffusion}  & PDM & U-Net& 256 & LSUN-Cat & \\
%       DiT-XL a~\cite{dit} & LDM & ViT & 256 & ImageNet & \\
%       DiT-XL b~\cite{dit}  & LDM & ViT & 512 & ImageNet & \\
%     \bottomrule
%     \end{tabular}
%     \caption{Models}
%     \label{tab:my_label}
% \end{table}

%%%%%%%%%%%%%%%%%%%%%%%%%%%%%%%%%%%%%%%%%%%%%%%%%%%%%%%%%%%%

\end{document}